\documentclass[letterpaper, 10 pt, conference]{ieeeconf}  

\IEEEoverridecommandlockouts                              

\usepackage{amsmath,amsfonts}
\usepackage{algorithm}
\usepackage{algpseudocode}
\usepackage{array}
\usepackage[caption=false,font=normalsize,labelfont=sf,textfont=sf]{subfig}
\usepackage{makecell}
\usepackage{hyperref}
\usepackage{amssymb}
\usepackage{svg}
\usepackage{gensymb}
\usepackage{textcomp}
\usepackage{booktabs}
\usepackage{stfloats}
\usepackage{url}
\usepackage{verbatim}
\usepackage{graphicx}
\usepackage{multirow}
\usepackage{soul}
\usepackage{newtxtext, newtxmath} 

\usepackage[dvipsnames]{xcolor}

\usepackage{dsfont}              
\newcommand{\ind}{\mathds{1}}

\usepackage{color}
\usepackage{capt-of}
\definecolor{highlight}{RGB}{205, 232, 248}

\newcommand{\first}[1]{\colorbox[HTML]{c0e2ca}{#1}}   
\newcommand{\second}[1]{\colorbox[HTML]{fff5b3}{#1}}  
\newcommand{\third}[1]{\colorbox[HTML]{ffd9b3}{#1}}   

\title{\textbf{CoPlanner: An Interactive Motion Planner with Contingency-Aware Diffusion for Autonomous Driving}}

\author{Ruiguo Zhong, Ruoyu Yao, Pei Liu, Xiaolong Chen, Rui Yang, Jun Ma, \textit{Senior Member, IEEE}
\thanks{Ruiguo Zhong, Ruoyu Yao, Pei Liu,  Rui Yang, Xiaolong Chen, and Jun Ma are with The Hong Kong University of Science and Technology (Guangzhou), Guangzhou 511453, China (e-mail: \{rzhong151, ryao092, pliu06l, ryang253\}@connect.hkust-gz.edu.cn;  xiaolongchen@hkust-gz.edu.cn; jun.ma@ust.hk).}
}

\begin{document}
\maketitle


\begin{abstract}
    Accurate trajectory prediction and motion planning are crucial for autonomous driving systems to navigate safely in complex, interactive environments characterized by multimodal uncertainties. However, current generation-then-evaluation frameworks typically construct multiple plausible trajectory hypotheses but ultimately adopt a single most likely outcome, leading to overconfident decisions and a lack of fallback strategies that are vital for safety in rare but critical scenarios. Moreover, the usual decoupling of prediction and planning modules could result in socially inconsistent or unrealistic joint trajectories, especially in highly interactive traffic. 
    To address these challenges, we propose a contingency-aware
diffusion planner (CoPlanner), a unified framework that jointly models multi-agent interactive trajectory generation and contingency-aware motion planning. 
Specifically, the pivot-conditioned diffusion mechanism anchors trajectory sampling on a validated, shared short-term segment to preserve temporal consistency, while stochastically generating diverse long-horizon branches that capture multimodal motion evolutions.
    In parallel, we design a contingency-aware multi-scenario scoring strategy that evaluates candidate ego trajectories across multiple plausible long-horizon evolution scenarios, balancing safety, progress, and comfort. 
    This integrated design preserves feasible fallback options and enhances robustness under uncertainty, leading to more realistic interaction-aware planning. 
    Extensive closed-loop experiments on the nuPlan benchmark demonstrate that CoPlanner consistently surpasses state-of-the-art methods on both Val14 and Test14 datasets, achieving significant improvements in safety and comfort under both reactive and non-reactive settings. Code and model will be made publicly available upon acceptance.
\end{abstract}

\section{Introduction}








Trajectory prediction and motion planning are fundamental components of autonomous driving, which enable the anticipation of surrounding agents' behaviors and facilitate informed decision-making to ensure safety and efficiency in complex driving scenarios \cite{hagedorn2024integration, yao2025hierarchical}. Recent advances in scene-level multi-agent trajectory generation, which models the joint distribution of all agents’ trajectories, demonstrate that diffusion models generate diverse, multimodal futures~\cite{huang2024versatile,yang2025wcdt}.
This generation step is typically followed by an evaluator that selects an executable plan, forming a generation-then-evaluation framework~\cite{Dauner2023CORL,huang2024gen}.
However, collapsing to a single most-likely future leaves the planner unprepared for alternative evolutions. When the scene unfolds differently, it cannot accommodate the broader set of plausible outcomes, leading to late braking, unstable execution, or unsafe maneuvers.



Contingency or risk-aware planning mitigates some of these issues by maintaining multiple candidate plans that share a short-term segment and branch later, deferring commitment until uncertainty resolves \cite{cui2021lookout, mustafa2024racp, bouzidi2025reachability}. 
However, in optimization-based formulations, predictions are usually treated as exogenous inputs and the planner optimizes trajectories conditioned on them. This decouples prediction from planning
 \cite{salzmann2020trajectron++, chen2024ir, liu2024reasoning}, which yields unrealistic or socially inconsistent joint trajectories, particularly in highly interactive or congested situations.


Recent generation-then-evaluation frameworks sample multiple futures and rank them with an evaluator \cite{Dauner2023CORL, huang2024gen}. In practice, the evaluator scores each scenario independently and then collapses to a single highest-scored one, discarding valuable fallback options and ignoring the mutual exclusivity of alternative behaviors for the same agent. In addition, without a shared short-term anchor, candidates often diverge too early, which harms execution stability and yields plans that are inconsistent under equally plausible evolutions. This mismatch makes it challenging for learning-based planners to naturally integrate contingency planning into their decision-making process.



Robust planning requires both preserving a consistent short-term execution and retaining feasible long-term trajectories for contingency. To this end, we propose a \textbf{Co}ntingency-Aware
Diffusion \textbf{Planner} (\textbf{CoPlanner}), a unified generation-then-evaluation framework that explicitly considers contingency with diffusion-based motion planning. A key innovation of our method is a pivot-conditioned diffusion mechanism, which anchors all sampled joint trajectories to a validated shared short-term segment while generating diverse long-term branches to capture multimodal futures.
To align short-term stability with long-term preparedness, we introduce a contingency-aware multi-scenario scoring rule that evaluates each ego candidate plan across multiple generated futures, balancing safety, progress, and comfort.
By coupling multimodal, interaction-aware generation with short-term segment consistency and multi-scenario evaluation, CoPlanner maintains feasible backup options and improves decision-making when uncertainty matters most.
Our contributions are summarized as follows:
\begin{itemize}

\item We introduce CoPlanner, a unified generation-then-evaluation framework that jointly samples ego plans and also the trajectories of other agents, preserving near-term consistency while keeping contingency branches. This ensures the ego vehicle remains feasible across plausible futures and enhances overall driving performance.

\item We propose a pivot-conditioned diffusion mechanism, a masked reverse-diffusion scheme that freezes a validated shared short-term segment and generates diverse long-term branches. This improves temporal consistency while preserving multimodality.
\item We develop a contingency‑aware scoring strategy that pre-filters candidates on a short-term horizon and then aggregates branch scores across multiple generated futures. This mitigates worst-case over-conservatism and selects plans that remain feasible, safe, and comfortable across scenarios, without sacrificing driving efficiency.
\item We conduct comprehensive closed-loop evaluations on nuPlan, achieving state-of-the-art performance on Val14 and Test14, with improved safety and comfort metrics.
\end{itemize}




\section{Related Work}

\begin{figure*}
    \centering
    \includegraphics[width=1\linewidth]{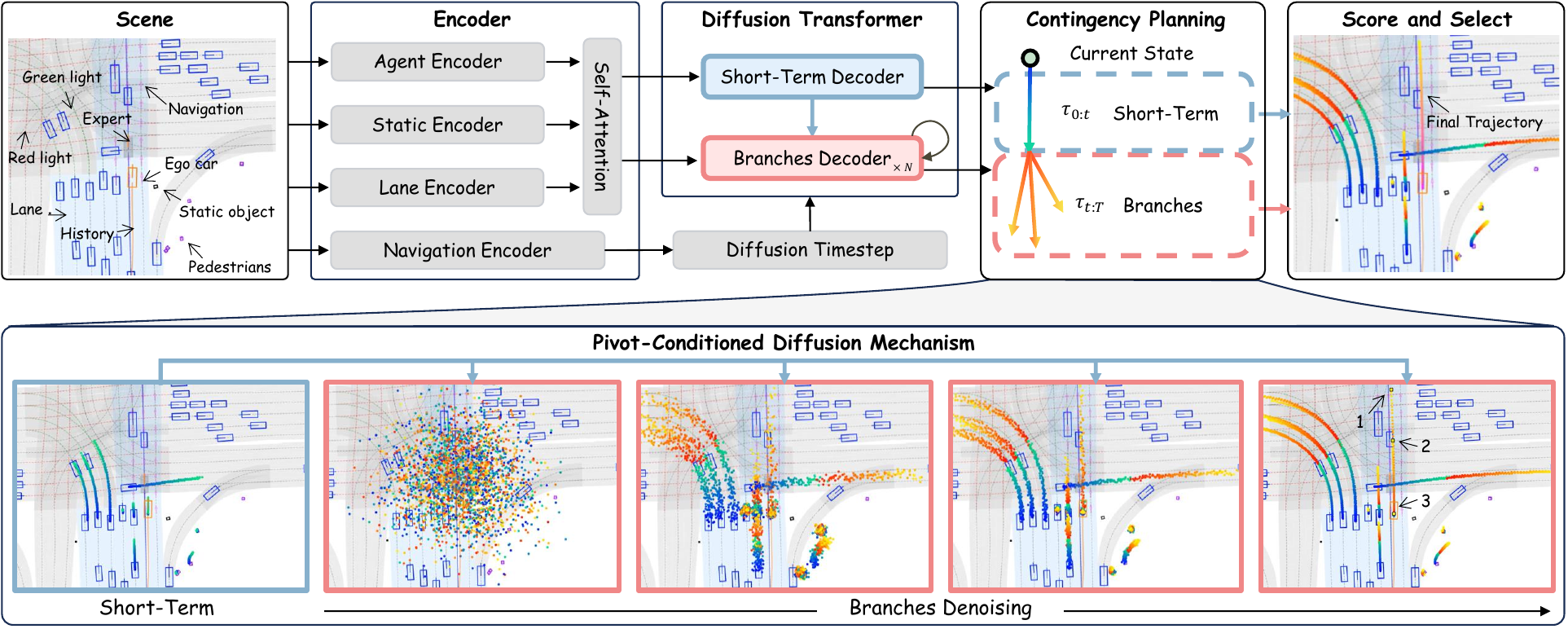}
    \caption{Overview of CoPlanner. From vectorized scene context, the planner first generates a set of shared short-term segments for all agents. Conditioned on these anchors, pivot-conditioned diffusion mechanism completes multiple diverse long-horizon branches via masked reverse denoising. A contingency-aware scorer evaluates across scenarios and selects the final ego trajectory.}

    \label{fig:pipeline}
\end{figure*}


\subsection{Multi-Modal Behavior Prediction}
Forecasting multiple plausible futures is essential for safe planning in interactive traffic. Vectorized scene representations that encode lanes, topology, and agent histories as polylines markedly improve the fidelity of learned predictors \cite{VectorNet_2020_CVPR, shi2024mtr++}. Prior multimodal methods emphasize per-agent diversity \cite{CoverNet_2020_CVPR, DESIRE_2017_CVPR}, but often struggle to produce coherent scene-level joint behaviors because inter-agent interactions are weakly modeled.
Generative approaches address this by sampling plausible joint futures one at a time, thereby capturing both multimodality and cross-agent dependencies. Diffusion-based predictors, in particular, demonstrate strong diversity and realism in trajectory generation \cite{huang2024versatile}. However, most such models remain downstream-agnostic: they generate agent futures without considering how a planner will exploit or contradict them. This disconnect can yield ego plans that appear feasible in isolation but are socially inconsistent once other agents react. Our work targets this gap by pairing multimodal generation with a mechanism that enforces a shared, safety-validated short-term segment across all candidates, making the predictions directly usable by a contingency-aware planner.

\subsection{Integrated Predictions and Planning}

Traditional autonomous driving stacks separate prediction and planning: forecasts are produced first and then treated as fixed inputs to the planner. This modularity simplifies engineering but can under-represent feedback between the ego vehicle and other agents in interactive scenes. To capture such coupling, recent research explores integrated formulations that allow mutual influence between prediction and planning \cite{hagedorn2024integration, sun2022m2i}. 
One strand models interaction with game theory, which iteratively updates each agent’s policy via best-response or equilibrium computations. Recent work addresses intent uncertainty and improves scalability by casting interactions as potential games or trajectory-tree–based pairwise games \cite{11081470, huang2025fast}. These methods are sensitive to intent priors and assumptions about other agents. 
A related line learns interaction patterns from data within a level-$k$ framework, including GameFormer \cite{huang2023gameformer} and HPP \cite{HPP}, while still relying on iterative rollouts. In practice, such iterative trajectory generation can yield jittery closed-loop behavior.
Beyond these, learning-based approaches treat interactions implicitly. 
Specifically, imitation-style planners like CALMMDrive \cite{CALMMDrive2025} generate diverse ego candidates and select the highest-scoring one under a vision–language metric.
Other approaches, such as Int2Planner \cite{chen2025int2planner}, add an explicit prediction head and score trajectories jointly, yet typically execute only the single top-confidence trajectory. Generative scene roll-outs, notably GenDrive \cite{huang2024gen}, further produce joint trajectories of ego vehicle and other agents and rank them with a vision–language model.
In practice, many integrated systems still collapse to a single most-likely outcome at decision time. This discards valuable fallback options, overlooks the mutual exclusivity of alternative behaviors for the same agent, and can cause early divergence among candidates, harming short-term executability. The approach proposed in this paper complements these lines by explicitly enforcing near-term consistency (a shared short-term segment) while retaining long-term diversity for contingency.

\subsection{Contingency Planning}
Unlike prior work that either focuses on multi-modal prediction, or contingency-aware planning without joint modeling, our method addresses the problem of consistent execution across scenarios by introducing a unified framework that jointly models a shared short-term segment and diverse long-term branches across futures \cite{zhenglei2025occlusion, zheng2025safe}.
CMPC \cite{bouzidi2025reachability} and RACP \cite{mustafa2024racp} synthesize a bundle of trajectories that share an initial safe segment before branching, thereby hedging against model error or perception noise. 
Sampling‑based approaches, such as LookOut \cite{cui2021lookout}, augment this idea with stochastic rollout and collision‑set pruning. While effective in low‑interaction scenes, these planners presume that every other agent follows a fixed, independently generated forecast.
Moreover, many contingency planners implicitly marginalize over futures, requiring the ego vehicle to trade low-probability but high-impact events against nominal progress without explicitly reasoning about mutually exclusive outcomes. Our framework couples contingency planning with scene-level joint trajectories generation for the ego vehicle and other agents by enforcing a shared short-term segment and sampling diverse long-term branches, which ensures feasibility across multiple plausible evolutions.

In summary, prior works suffer from three main limitations: (i) multimodal predictors generate diverse, coherent joint futures but are decoupled from planning, with no mechanism for using those samples in decision making; (ii) integrated predictions and planning systems often collapse to a single most-likely hypothesis at decision time, leaving no fallback when reality deviates and causing policy chattering; and (iii) contingency planners keep a shared prefix with late branching but typically optimize against exogenous, fixed forecasts, which weakens social consistency. 
We bridge these gaps by combining short-term conditioned diffusion with multi-scenario evaluation, yielding plans that are both stable to execute and robust under uncertainty.


\section{Problem Formulation}


We consider a driving scene with $M$ agents indexed by $j\in\{0,1,\dots,M{-}1\}$, where $j{=}0$ denotes the ego vehicle and $j\ge 1$ denote the other agents. 
Time is discretized as $\{0,1,\dots,T\}$ with step $\Delta t$.
The future trajectory of agent $j$ is $\tau^j=\{x^j_t\}_{t=0}^T$, and the joint future is
$\Lambda \!=\! \{\tau^j\}_{j=0}^{M-1}$.
Context $\boldsymbol{C}$ includes agent histories, HD map, traffic signals, and route.

The classical objective of autonomous driving is to seek one ego plan $\tau^0\!\in\!\mathcal T$ minimizing a nominal cost $J$ under feasibility constraints:
\begin{align}
    \tau^* &= \arg \min_{\substack{\tau \in \mathcal{T}}} J(\tau)
\end{align}
where $\mathcal{T}$ is the set of feasible candidate trajectories given the current ego state.  

This single-scenario view implicitly commits to one most-likely evolution of the environment. To explicitly consider contingency, we introduce a branching time $t_b\!\in\!\{0,1,\dots,T\}$ and split the joint future into a shared short-term segment $ S=\{\tau^j_{0:t_b}\}_{j=0}^{M-1}$ and a long-term branch $ B=\{\tau^j_{t_b:T}\}_{j=0}^{M-1}$ so that $\Lambda=[ S, B]$. For an ego trajectory $\tau^0$ and some potential future $\Lambda$, we decompose the evaluation window:
\begin{equation}
J(\tau^0\mid \Lambda, \boldsymbol{C})=
J_{\text{shared}}\big(\tau^0_{0:t_b}\mid \Lambda, \boldsymbol{C}\big)+
J_{\text{branch}}\big(\tau^0_{t_b:T}\mid \Lambda, \boldsymbol{C}\big).
\end{equation}
A contingency policy executes an identical short-term ego trajectory segment under all plausible evolutions and branches later. Denote the ego anchor by $\hat\tau^0_{0:t_b}$ and the feasible set with that anchor by
\begin{equation}
\mathcal T_{\text{short}}(\hat\tau^0_{0:t_b})
=\big\{\tau^0\in\mathcal T:\;\tau^0_{0:t_b}=\hat\tau^0_{0:t_b}\big\}.
\end{equation}




Rather than treating $\Lambda$ as exogenous \cite{mustafa2024racp}, we model the joint future generatively by factorizing it into a shared short-term segment $S$ and a long-term branch $B$:
\begin{equation}
p(\Lambda\mid \boldsymbol{C})
= p_\theta( S\mid \boldsymbol{C})\;p_\phi( B\mid  S,\boldsymbol{C}).
\label{eq:gen-factor}
\end{equation}
\noindent Here, $\theta$ and $\phi$ parameterize the short-term generator and the conditional long-term completion model, respectively. At test time we sample $K$ short-term segments and, for each segment, sample $N$ long-term branches:
\begin{equation}
\hat S_k \sim p_{\theta}(S \mid \boldsymbol{C}),\ 
B_{k,n} \sim p_{\phi}(B \mid \hat S_k, \boldsymbol{C}),\ 
\Lambda_{k,n} = [\hat S_k, B_{k,n}],
\end{equation}
with $k=1,2,\dots,K$ and $n=1,2,\dots,N$. 


For a retained short-term segment $\hat S_k$ we select one ego plan that shares $\hat\tau^{0,k}_{0:t_b}$ and is robust across its $N$ plausible futures:
\begin{align}
J_k(\tau^0) &= J_{\text{shared}}\!\big(\tau^0_{0:t_b}\mid \hat S_k,\boldsymbol{C}\big) \\ \notag
& +~\mathcal R\!\left(\left\{J_{\text{branch}}\!\big(\tau^0_{t_b:T}\mid \Lambda_{k,n},\boldsymbol{C}\big)\right\}_{n=1}^{N}\right), \\
\tilde\tau^{0,k}
&= \arg\min_{\tau^0\in \mathcal T_{\text{short}}(\hat\tau^{0,k}_{0:t_b})} J_k(\tau^0).
\label{eq:inner-opt}
\end{align}

In this work, we instantiate the risk aggregator $\mathcal R$ as the equal-weight mean over the generated futures ${\Lambda_{k,n}}$.

\section{Methodology}

\subsection{Overview} \label{sec:overview}



An overview of the full pipeline is shown in Fig.~\ref{fig:pipeline}. CoPlanner executes a generation-then-evaluation loop at each cycle:
(i) \textbf{Short-term segment generation} produces $K$ shared short-term segments $\{\hat S_k\}$ and keeps those passing hard checks on $[0,t_b]$;
(ii) for each retained short-term segment, \textbf{pivot-conditioned diffusion mechanism} samples $N$ diverse long-term branches $\{ B_{k,n}\}$ to form $\{\Lambda_{k,n}\}$ that all share the same near-term joint evolution;
(iii) \textbf{two-stage contingency scoring} first pre-filters ego candidates on $[0,t_b]$, then aggregates branch costs over $\{\Lambda_{k,n}\}$ on $(t_b,T]$ to obtain $\tilde\tau^{0,k}$; 
(iv) a selection across short-term segment groups outputs the final $\tau^{0,*}$. This design separates near-term stability from long-term preparedness while keeping them consistent.

\begin{figure}
    \centering
    \includegraphics[width=1\linewidth]{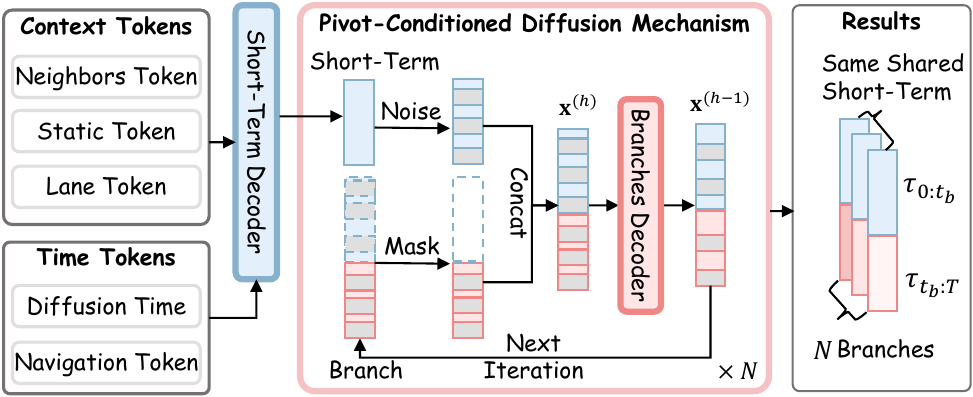}
    \caption{Pivot-conditioned diffusion mechanism with masked reverse diffusion.
    At each reverse step, short-term frames ($t\le t_b$) are re-noised from the fixed anchor and overwrite the state,
while long-term frames ($t>t_b$) are denoised by $G_{\text{full}}$ and merged via the mask.
    This guarantees a shared near-term execution and late branching.}
    \label{fig:denoising}
\end{figure}

\subsection{Diffusion-based Scenario Generator}
\label{sec:scen-gen}
\noindent \textbf{Backbone and Objective.} We adopt a Diffusion Transformer (DiT) backbone operating on stacked scene trajectories for the ego vehicle and other agents. 
We use a variance-preserving (VP) diffusion process with continuous time $h\!\in\![0,1]$ that is discretized into $H$ steps at inference. 
At each diffusion step $h$, the noised future $\mathbf{x}^{(h)}$ is fused with the scene context $\boldsymbol{C}$ via gated cross-attention and a sinusoidal timestep embedding. To improve stability under heavy conditioning, we use the $\mathbf{x}^{(0)}$-prediction objective, training the denoiser to directly reconstruct the clean sample:
\begin{equation}
\mathcal{L}_\theta
= \mathbb{E}_{\mathbf{x}^{(0)},\, h\sim\mathcal{U}(0,1),\, \mathbf{x}^{(h)} \sim q_{h|0}}
\Big[\,\big\|\mu_\theta(\mathbf{x}^{(h)}, h, \boldsymbol{C}) - \mathbf{x}^{(0)}\big\|_2^2\,\Big].
\end{equation}
For sampling, the continuous VP schedule is discretized to $\{\alpha_h\}_{h=1}^{H}$ with $\bar\alpha_h=\prod_{s=1}^{h}\alpha_s$, which matches the continuous schedule at the corresponding time, ensuring train–test consistency.

\noindent \textbf{Short-Term Model} $G_{\text{shared}}$. $G_{\text{shared}}$ 
models the distribution of shared short-horizon joint futures $p_\theta( S\mid \boldsymbol{C})$ and is trained to produce high-confidence short-term segments over the interval $[0,t_b]$. The input is the scene context $\boldsymbol{C}$ containing agent histories, a vectorized map, traffic controls, and the route. It uses the same diffusion backbone and the $\mathbf{x}^{(0)}$-prediction objective as the full model, with the loss applied only on the short-term window.

\noindent \textbf{Full Model} $G_{\text{full}}$. $G_{\text{full}}$ models the conditional distribution of long-term branches, $p_\phi(B\mid S, \boldsymbol{C})$, and completes the scene evolution given a fixed shared short-term segment. Its inputs are the scene context $\boldsymbol{C}$ and a clean anchor short-term segment $ S$, and its output is the joint trajectory on $[0,T]$. 
Training only on the long-term portion deprives the model of exposure to early-phase dynamics and interactions, which leads to mode collapse, boundary artifacts around $t_b$, and violations of vehicle kinematics. 
To avoid this, we predict the full horizon while applying a long-term corruption mask that injects noise only for $t>t_b$. The $\mathbf{x}^{(0)}$-prediction objective is defined on $[0,T]$, whereas the diffusion loss is evaluated on branch frames. The clean short-term segment is provided as a conditioning signal, with optional short-term conditioning dropout to prevent over-reliance.

\subsection{Pivot-Conditioned Diffusion Mechanism}
\label{sec:PCDM}

We apply a binary temporal mask $m(t)=\ind[t\le t_b]$ during sampling, where $\ind\{\cdot\}$ denotes the indicator; short-term frames ($t\!\le\!t_b$) are frozen and long-term frames ($t\!>\!t_b$) are denoised.
The operator $\mathrm{re\mbox{-}noise}(\hat S)$ re-applies the forward noising kernel to the clean short-term anchor $\hat S$ at the current reverse step, i.e., it draws $x^{(h-1)}_{\text{short}}\sim q_{h-1|0}(\cdot\,|\,\hat S)$ and overwrites the short-term entries of the state.

\noindent \textbf{Step A: Multiple Short-Term Segments Sampling and Pre-Filtering.}
In this part, we first sample $K$ joint short-term segments $\{\hat S_k\}\!\sim\!p_\theta( S\mid\boldsymbol{C})$.
For each candidate, we compute a PDM score over the short-term window $[0,t_b]$ that evaluates safety, rule compliance, comfort, and progress. 
If all scores are zero, the scene is deemed unsafe for nominal execution and an emergency-braking profile is triggered for this cycle.
Otherwise, the remaining short-term segments are taken as anchors for the next steps.

\noindent\textbf{Step B: Multi-Branch Completion Under Each Short-Term Segment.}
In this part, we complete diverse long-term branches conditioned on each retained shared short-term segment, with the goal of keeping identical near-term execution while exploring multiple plausible futures.
For a retained short-term segment $\hat S_k$, we run masked reverse diffusion to generate the long-term branch while freezing all short-term frames, yielding $N$ futures $\Lambda_{k,n}=[\hat S_k, B_{k,n}]$. 
The full sampling procedure is summarized in Algorithm~\ref{alg:PCDM}, where the noise term $\epsilon_\phi$ is derived from the model’s $\mathbf{x}^{(0)}$ estimate via the standard variance-preserving (VP) conversion; we set $\beta_h=1-\alpha_h$ and choose $\sigma_h$ according to the VP schedule. At each reverse step, known short-term segments are re-noised from the anchor and overwrite the corresponding states; unknown branch frames are denoised by $G_{\text{full}}$ and merged by the mask. This guarantees that all $\Lambda_{k,n}$ share the same near-term execution and branch only on $(t_b,T]$. As illustrated in Fig.~\ref{fig:denoising}, the known short-term segment is repeatedly re-noised to enforce consistency, and the unknown branch is generated by the model.

\begin{algorithm}[t]
\caption{Pivot-Conditioned Diffusion Mechanism}
\label{alg:PCDM}
\small
\begin{algorithmic}[1]
\State \textbf{Input:} Context $\boldsymbol{C}$, Short-Term Segment $\hat S$, Mask $m(t)=\ind[t\le t_b]$, steps $H$
\For{$n=1$ to $N$} \Comment{Sample $N$ branches}
  \State $\mathbf{x}^{(H)}\!\sim\!\mathcal N(0,I)$
  \For{$h=H$ \textbf{to} $1$}
    \State $z\!\sim\!\mathcal N(0,I)$ if $h{>}1$ else $z{=}0$
    \State $\mathbf{x}^{(h-1)}_{\text{short}}\!\leftarrow\!\text{re-noise}\ (\hat S)$ \Comment{Freeze short-term segment}
    \State $\mathbf{x}^{(h-1)}_{\text{branch}}\!\leftarrow\!\frac{1}{\sqrt{\alpha_h}}\!\Big(\mathbf{x}^{(h)}-\frac{\beta_h}{\sqrt{1-\bar\alpha_h}}\,
    \epsilon_\phi(\mathbf{x}^{(h)},h,\boldsymbol{C},\hat S)\Big)+\sigma_h z$
    \State $\mathbf{x}^{(h-1)} \leftarrow m\odot \mathbf{x}^{(h-1)}_{\text{short}} + (1{-}m)\odot \mathbf{x}^{(h-1)}_{\text{branch}}$
  \EndFor
  \State $ B_{k,n}\leftarrow \mathbf{x}^{(0)}_{t_b:T}$,\quad $\Lambda_{k,n} \leftarrow [\hat S, B_{k,n}]$
\EndFor
\State \textbf{return} $\{\Lambda_{k,n}\}_{n=1}^{N}$
\end{algorithmic}
\end{algorithm}

\noindent \textbf{Step C: Ego Candidate Set Under a Shared Short-Term Segment.}
For each retained shared short-term segment $\hat S_k$, we build a separate ego candidate set $\mathcal T_{\text{short}}^{(k)}$. Within each set, all candidates are identical to the group’s anchor $\hat\tau^{0,k}_{0:t_b}$ on $[0,t_b]$ and branch only on $(t_b,T]$. Candidates are obtained exclusively by reusing the ego branches produced by $G_{\text{full}}$ under that short-term segment, yielding $N$ alternative futures. The concrete scoring and selection procedure is described in Sec.~\ref{sec:scoring}.

\subsection{Two-Stage Contingency-Aware Scoring}
\label{sec:scoring}

\noindent\textbf{Stage~1: Short-Term Segment Pre-Filter ($[0,t_b]$).}
Within each short-term–segment group, all ego candidates share the same anchor on $[0,t_b]$; hence, the short-term score is a group-wise constant. For every retained short-term segment $\hat{S}_k$ we compute
\begin{equation}
E^{\mathrm{pre}}_k
\;=\;
\mathrm{PDM}_{t_b}\!\big(\hat{\tau}^{0,k}_{0:t_b}\,\big|\,\hat{ S}_k,\boldsymbol{C}\big).
\end{equation}
We retain only groups with $E^{\mathrm{pre}}_k>0$. If $\max_k E^{\mathrm{pre}}_k=0$, no short-term segment satisfies the executability criterion, and an emergency braking profile is issued for the current cycle.

\medskip
\noindent\textbf{Stage~2: Long-Term Window Multi-Scenario Aggregation.}
For each surviving short-term segment $k$ and any ego candidate $\tau^0\in\mathcal{T}_{\text{short}}^{(k)}$, we evaluate the full-horizon PDM score on each generated future $\Lambda_{k,n}$:
\begin{equation}
E_{k,n}(\tau^0)
\;=\;
\mathrm{PDM}_{T}\!\big(\tau^0 \,\big|\, \Lambda_{k,n},\boldsymbol{C}\big),
\quad n=1,2,\ldots,N,
\end{equation}
where $\mathrm{PDM}_{T}$ denotes the PDM score over the full planning horizon $[0,T]$; in our experiments we instantiate it with $T=8\,\mathrm{s}$ (i.e., $\mathrm{PDM}_{8\mathrm{s}}$). We aggregate across scenarios by the mean
\begin{equation}
\bar{E}_k(\tau^0)
\;=\;
\frac{1}{N}\sum_{n=1}^{N} E_{k,n}(\tau^0).
\end{equation}
The best candidate within the short-term segment $k$ is then selected as
\begin{equation}
\tilde{\tau}^{0,k}
\;=\;
\arg\max_{\tau^0\in\mathcal{T}_{\text{short}}^{(k)}} \; \bar{E}_k(\tau^0),
\end{equation}
followed by cross-group selection across the short-term segments:
\begin{equation}
k^*
\;=\;
\arg\max_{k} \; \bar{E}_k\!\big(\tilde{\tau}^{0,k}\big),
\qquad
\tau^{0,*}
\;=\;
\tilde{\tau}^{0,k^*}.
\end{equation}
 Stage~1 serves as an executability pre-filter, whereas Stage~2 selects the within-group and global optima by the mean $\mathrm{PDM}_{T}$ across generated futures.

\section{Experiments}

\subsection{Experimental Setup}
\begin{table*}[t]
    \centering
    \caption{\small Closed-loop planning results on nuPlan dataset. Columns are ordered as Test14-hard, Test14, then Val14. Colors denote the top-3 \textbf{within each group} (Learning-based vs. Rule-based \& Hybrid): \first{first}, \second{second}, \third{third}. \textbf{Bold} = overall best per column (excluding Expert). NR: non-reactive; R: reactive.}
    \label{tab:nuplan}
    \vspace{-5pt}
    \resizebox{0.8\linewidth}{!}{\scriptsize
    \begin{tabular}{llcccccc}
    \toprule
    \multirow{2}{*}[-0.15ex]{\makecell[l]{\textbf{Type}}} & \multirow{2}{*}[-0.15ex]{\makecell[l]{\textbf{Planner}}} & \multicolumn{2}{c}{\textbf{Test14-hard}} & \multicolumn{2}{c}{\textbf{Test14}} & \multicolumn{2}{c}{\textbf{Val14}} \\
    \cmidrule(lr){3-4} \cmidrule(lr){5-6} \cmidrule(lr){7-8}
    & & \textbf{NR} & \textbf{R} & \textbf{NR} & \textbf{R} & \textbf{NR} & \textbf{R} \\ \midrule

    \textcolor{gray}{Expert} & \textcolor{gray}{Log-replay} & \textcolor{gray}{85.96} & \textcolor{gray}{68.80} & \textcolor{gray}{94.03} & \textcolor{gray}{75.86} & \textcolor{gray}{93.53} & \textcolor{gray}{80.32} \\ \midrule

    \multirow{6}{*}[0.5ex]{\makecell[l]{Learning-based}}
    & PDM-Open                & 33.51 & 35.83 & 52.81 & 57.23 & 53.53 & 54.24 \\
    & UrbanDriver                       & 50.40 & 49.95 & 51.83 & 67.15 & 68.57 & 64.11 \\
    & PlanTF                            & 69.70 & \second{61.61} & 85.62 & \first{79.58} & 84.27 & \third{76.95} \\
    & PLUTO w/o refine      & \third{70.03} & 59.74 & \second{89.90} & \third{78.62} & \second{88.89} & \second{78.11} \\
    & Diffusion Planner w/o refine      & \second{72.39} & \third{61.21} & \third{87.33} & 75.51 & \third{86.87} & 75.30 \\
    & CoPlanner w/o refine \textbf{(Ours)}    & \first{76.82} & \first{64.47} & \first{90.31} & \second{78.81} & \first{89.48} & \first{79.00} \\
    \midrule

    \multirow{7}{*}[-0.15ex]{\makecell[l]{Rule-based \\ \& Hybrid}}
    & IDM                               & 56.15 & 62.26 & 70.39 & 74.42 & 75.60 & 77.33 \\
    & PDM-Closed                        & 65.08 & 75.19 & 90.05 & \third{91.63} & 92.84 & \third{92.12} \\
    & PDM-Hybrid                        & 65.99 & 76.07 & 90.10 & 91.28 & 92.77 & 92.11 \\
    & GameFormer                        & 68.70 & 67.05 & 83.88 & 82.05 & 79.94 & 79.78 \\
    & PLUTO                              & \first{\textbf{80.08}} & \third{76.88} & \third{92.23} & 90.29 & \third{92.88} & 76.88 \\
    & Diffusion Planner w/ refine & \second{78.87} & \first{\textbf{82.00}} & \first{\textbf{94.80}} & \second{91.75} & \second{94.26} & \second{92.90} \\
    & CoPlanner w/ refine \textbf{(Ours)}     & \third{76.05} & \second{78.59} & \second{93.47} & \first{\textbf{92.00}} & \first{\textbf{94.45}} & \first{\textbf{93.13}} \\
    \bottomrule
    \end{tabular}}
\end{table*}

\noindent \textbf{Benchmark.} We evaluate our method on the nuPlan benchmark \cite{nuPlan2024benchmark}, a large-scale dataset for closed-loop motion planning. It provides $\sim$1{,}500 hours of expert driving across four cities and a simulator that initializes from logged scenarios and runs for $15\,\mathrm{s}$ at 10\,Hz. We use the official simulator, which initializes scenarios from real-world logs. Other agents are controlled by either log-replay (non-reactive, \textbf{NR}) or an IDM policy (reactive, \textbf{R}). Low-level tracking is handled by an LQR controller that converts the planned trajectory into throttle/brake/steer commands.  
We report results on 
\textbf{Test14-hard} (272 scenarios) \cite{cheng2024rethinking}, \textbf{Test14} (261 scenarios), and \textbf{Val14} (1118 scenarios),
each in both non-reactive and reactive modes.

\noindent \textbf{Baselines and Metrics.}
We compare CoPlanner against the following representative planners: IDM \cite{helbing1998generalized}; PDM* (including PDM-Closed, PDM-Hybrid, and PDM-Open) \cite{Dauner2023CORL}; UrbanDriver \cite{scheel2022urban}; GameFormer \cite{huang2023gameformer}; PlanTF \cite{cheng2024rethinking}; Diffusion Planner \cite{zheng2025diffusionbased}; and PLUTO \cite{cheng2024pluto} (reported both with and without refine). Unless otherwise noted, we use official implementations and public settings. For the Diffusion Planner (learning-only) variant, we retrain on our unified $\sim$1M nuPlan split. For the variants with post-processing (refine), we report the publicly released results without modification. Following nuPlan, we report the overall \textbf{nuPlan score} (0–100, higher is better) averaged over scenarios, together with key sub-metrics: Collisions, Time-to-collision (TTC), Drivable area compliance, Comfort, and Progress.

\subsection{Implementation Details}
\noindent \textbf{Data and Preprocessing.}
We train on $\sim$1M nuPlan clips with $2\,\mathrm{s}$ history and an $8\,\mathrm{s}$ planning horizon, sampled at 10\,Hz. Poses are expressed in an ego–centric frame at $t{=}0$ with per-scene normalization. For each scene, we keep the ego vehicle and the 32 nearest other agents within a 100\,m radius for context, while predicting only the 10 nearest agents. Agent histories include position and heading encoded as $(x,y,\cos\psi,\sin\psi)$. We also gather vectorized map and traffic-control elements and the route polyline within 100\,m of the ego vehicle at the current time; neighboring agents’ $2\,\mathrm{s}$ histories are stacked at 10\,Hz. All modalities are padded to fixed token slots, and attention masks prevent attending to padded tokens or invalid segments. Following the time horizon used in PDM’s hybrid setup and scoring scheme, we set the branching time to $t_b{=}4\,\mathrm{s}$.

\noindent\textbf{Models.}
We use two DiT models operating on stacked multi-agent trajectory tokens with gated cross-attention to context tokens (vectorized lanes/traffic controls, route, and agent histories) and sinusoidal timestep embeddings. 
The short-term model $G_{\text{shared}}$ generates a high-confidence shared short-horizon segment of length $t_b{=}4\,\mathrm{s}$; the full model $G_{\text{full}}$ performs pivot-conditioned diffusion mechanism to produce diverse long-term branches conditioned on that short-term segment. 
Both models are trained with an $\mathbf{x}^{(0)}$-prediction objective. For $G_{\text{full}}$, Gaussian noise is applied only to frames $t{>}t_b$ while short-term segment frames remain clean and are provided as conditions. To expose the model to full-trajectory multimodality, the $x_0$ target covers the entire $8\,\mathrm{s}$ horizon, and we apply short-term-conditioning dropout with a rate of $0.5$ during training (randomly masking the short-term segment condition to reduce over-reliance).
For sampling, we use DPM-Solver++ \cite{lu2022dpm} with a VP noise schedule. We train for 300 epochs with a global batch size of 512 on 4×RTX 4090 GPUs (about 2 days). The learning rate is initialized at $5\times10^{-4}$ and uses a 5-epoch warmup.

\noindent \textbf{CoPlanner.}
We implement CoPlanner on the nuPlan closed-loop benchmark. 
For candidate short-term segment generation, we retain $K=3$ short-term joint trajectories $\{\hat S_k\}_{k=1}^K$, each validated by the PDM score. If all scores are zero, we trigger an emergency-braking profile.
For each retained short-term segment $\hat S_k$, the pivot-conditioned diffusion mechanism samples $N=3$ diverse long-term branches. 
Ego planning results are projected to a kinematic bicycle model to ensure trackability; a low-level LQR converts the selected trajectory into control commands.

\section{Results and Discussion}
 All methods are evaluated in the official nuPlan closed-loop simulator in both non-reactive (log-replay) and reactive (IDM) settings with identical LQR tracking and controller limits to ensure fairness. To align with hybrid baselines, we attach a refinement module \cite{Dauner2023CORL, sun2024mixture} as the postprocessor to our approach (\textbf{CoPlanner w/ refine}).

\subsection{Main Results}
Table~\ref{tab:nuplan} summarizes closed-loop performance across \textbf{Test14-hard}, \textbf{Test14}, and \textbf{Val14} in both non-reactive (NR) and reactive (R) modes. On \textbf{Val14}, \textbf{CoPlanner w/ refine} achieves the best overall results in both NR and R; on \textbf{Test14}, \textbf{Diffusion Planner w/ refine} ranks first in NR while \textbf{CoPlanner w/ refine} ranks first in R and remains highly competitive in NR; on \textbf{Test14-hard}, \textbf{Diffusion Planner w/ refine} leads in R and is second in NR, with \textbf{CoPlanner w/ refine} close to the strongest hybrid baseline in NR. Notably, \textbf{CoPlanner w/o refine} already outperforms most learning-only baselines across splits, and the lightweight refinement consistently adds further gains. In this part, \textbf{CoPlanner w/o refine} and the single-stage Diffusion Planner share the same DiT backbone and training recipe; CoPlanner’s two-stage factorization lowers conditional entropy and simplifies credit assignment via shorter horizons, and reduces imitation bias.
Overall, these results validate the benefit of CoPlanner’s shared short-term segment generation and multi-scenario scoring: it improves safety and comfort in interactive scenarios without sacrificing progress and remains robust on tail-risk scenarios. 
We observe comparatively smaller gains on \textbf{Test14-hard}, which we attribute to our deliberate reliance on a PDM-style composite score for pre-filtering and ranking: this split is curated around scenarios where PDM scoring is known to be less informative, reducing the advantage of our scorer.

\subsection{Qualitative Results}
\begin{figure}
    \centering
    \includegraphics[width=1\linewidth]{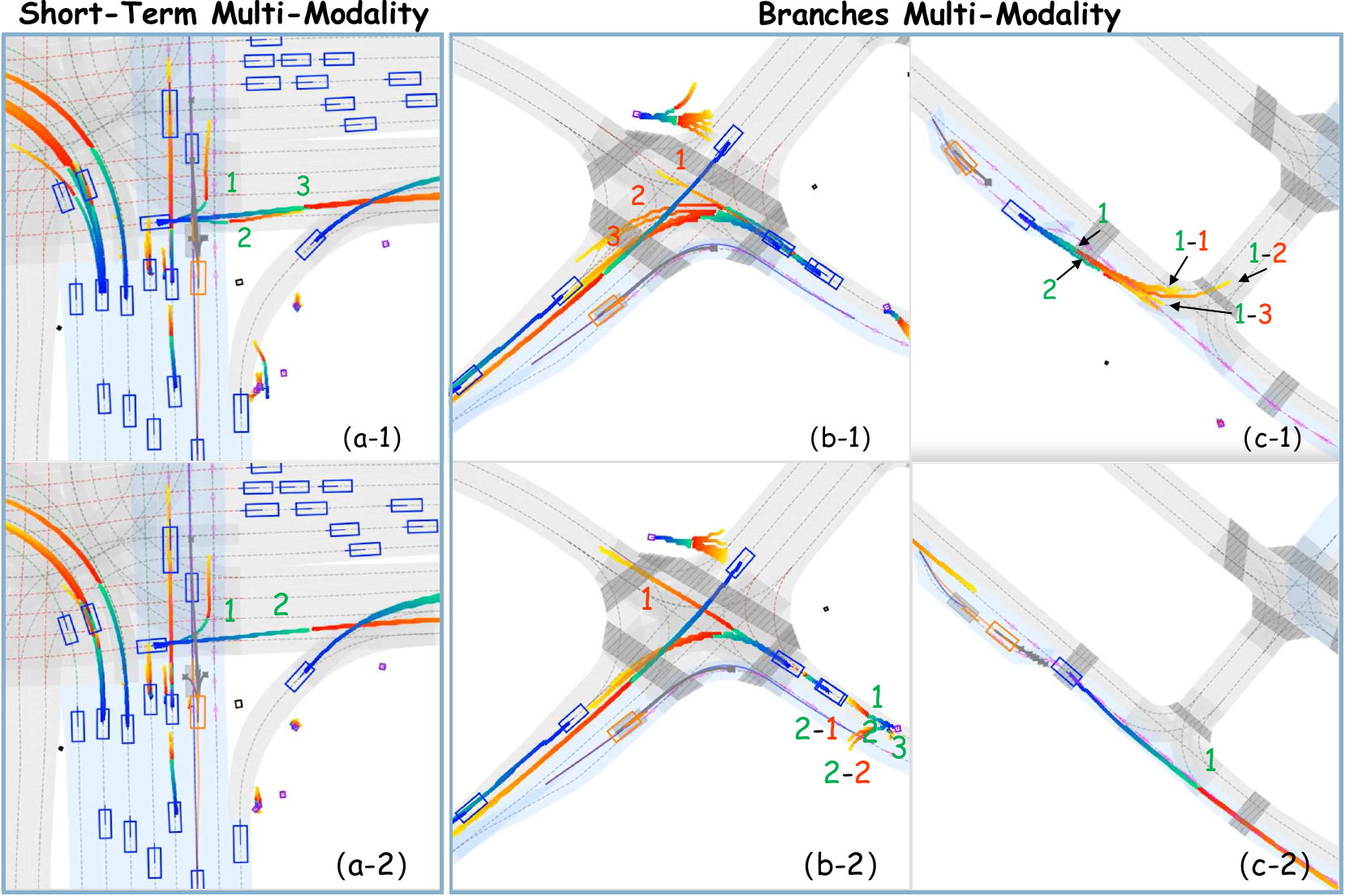}
    \caption{
    Top: under a shared \textcolor{ForestGreen}{short-term segment} (green), multiple \textcolor{BrickRed}{long-term branches} (red) emerge.
    Bottom: as time elapses, branches are pruned and converge to one or a few dominant outcomes.}
    \label{fig:multi_modal_3fig}
\end{figure}

Figure~\ref{fig:multi_modal_3fig} illustrates CoPlanner’s short-term conditioned behavior across three interaction scenes. In the top row, trajectories share a common green short-term segment and then branch into multiple red branches, capturing early-stage multimodality at merges, lane changes, and unprotected turns. In the bottom row, as time elapses and interactions resolve, many less likely branches are pruned by right of way, map topology, and social constraints, and the set converges to one or a few dominant outcomes. The shared short-term segment yields stable near-term execution and prevents plan switching, while the surviving branches keep feasible contingency options if surrounding agents brake, force their way in, or yield unpredictably. This leads to smoother and safer motion without sacrificing route progress.





\subsection{Ablation Studies}

\begin{figure}[t]
  \centering
  \includegraphics[width=1\linewidth]{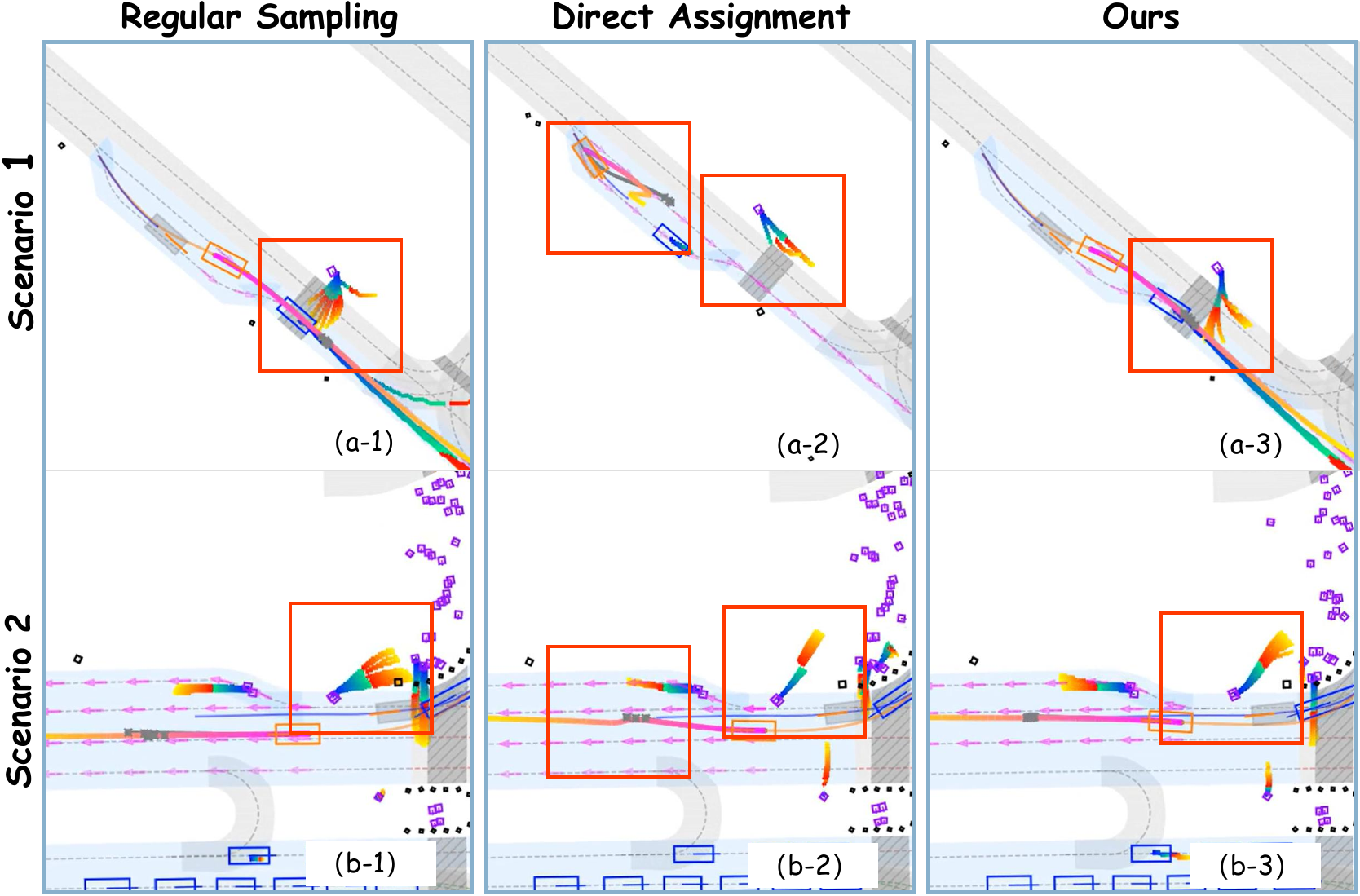}
  \caption{Comparison of three generation strategies for multi-modal trajectory generation.
  (a) Regular Sampling: multiple futures are generated independently per step, leading to scattered modes and temporal inconsistency.
  (b) Frame-wise short-term segment direct assignment: re-assigns the short-term segment at every frame, which reduces spread but causes jitter and mode flipping.
  (c) Ours: short-term segment consistent multi-modal prediction with late-stage branching, producing a stable shared short-term segment and diverse yet feasible terminations with improved temporal consistency.}
  \label{three_methods}
\end{figure}

\begin{table*}[t]
\centering
\vspace{3pt}
\caption{\small 
 Ablations on \textbf{nuPlan Val14 (Reactive)}. Higher is better.}
\vspace{-5pt}
\resizebox{1.\linewidth}{!}{\scriptsize
\begin{tabular}{llccccccc}
\toprule
\textbf{Block} & \textbf{Variant} & \textbf{Eval. Time} & \textbf{Score} & \textbf{Collisions} & \textbf{TTC} & \textbf{Drivable} & \textbf{Comfort} & \textbf{Progress} \\
\midrule
\multirow{2}{*}{\shortstack[l]{Planner Arch. \\ w/o refine}} & Diffusion Planner & - & 75.30 & 95.57 & 91.50 & 96.69 & 85.50 & 89.17 \\
& CoPlanner & - & 79.00 & 96.14 & 92.27 & 97.66 & 88.14 & 90.84 \\
\midrule
\multirow{2}{*}{Evaluation Horizon} & PDM Score & 8s & 91.97 & 98.73 & 95.20 & 99.64 & 89.04 & 98.37 \\
& PDM Score & 4s & 92.95 & 98.21 & 93.38 & 99.73 & 92.67 & 99.64 \\
\midrule
\multirow{3}{*}{\shortstack[l]{Generation Strategy \\ (Contingency Score)}} & Regular Sampling & 8s & 92.89 & 98.26 & 93.38 & 99.73 & 91.68 & 99.64 \\
& Direct Assignment & 8s & 93.03 & 98.35 & 94.63 & 99.82 & 94.19 & 99.02 \\
& Ours & 8s & \textbf{93.13} & 98.26 & \textbf{94.81} & 99.73 & \textbf{94.28} & 99.19 \\
\bottomrule
\end{tabular}}
\label{tab:ablation}
\end{table*}
We structure the ablations into three parts, summarized in Table~\ref{tab:ablation} and illustrated in Fig.~\ref{three_methods}: 
(i) architecture compares single-stage diffusion with our two-stage generation without refinement; 
(ii) scoring baseline keeps the refinement module and extends the PDM evaluation window to $8\,\mathrm{s}$; 
(iii) our method: Contingency Score with the same refinement and $8\,\mathrm{s}$ window, comparing three sampling strategies.

\noindent\textbf{Planner Architecture.}
In this experiment, all models are trained on the same dataset.  The two-stage design with a shared short-term segment followed by branch completion improves the nuPlan score by 3.7 points over the single-stage baseline and raises every submetric. This suggests that anchoring a validated shared short-term segment stabilizes near-term execution and yields more socially consistent interactions. Methodologically, decoupling the problem into a short-horizon executor and a short-term segment conditioned long-term completer reduces the effective learning horizon faced by imitation learning, mitigates imitation shift, and results in stronger closed-loop behavior.

\noindent\textbf{Evaluation Horizon.}
With the same refinement module, extending the PDM evaluation window from $4\,\mathrm{s}$ to $8\,\mathrm{s}$ lowers the overall nuPlan score. Safety metrics improve (fewer collisions, higher TTC), but comfort degrades noticeably and progress shows no gain. The longer window overemphasizes distant, hypothetical interactions and underweights near-term executability, prompting overly conservative behavior (earlier braking and jerkier control). This observation motivates our design: score the shared short-term segment on $[0,t_b]$ and aggregate long-term costs across multiple futures rather than stretching a single-window PDM.

\noindent\textbf{Generation Strategy Under Contingency Score.}
With the same refinement and $8\,\mathrm{s}$ window, Contingency Score combined with pivot-conditioned diffusion mechanism performs best among the three strategies considered.
Internally, we compare three sampling schemes in Fig.~\ref{three_methods}:  
(i) Regular sampling increases diversity but scatters modes and weakens temporal coherence;  
(ii) Frame-wise short-term segment direct assignment reduces dispersion but causes jitter and mode flips due to frequent re-anchoring;  
(iii) Pivot-conditioned diffusion mechanism maintains a consistent shared short-term segment before branching, balancing stability and diversity, and achieves the best overall score.
In particular, pivot-conditioned diffusion mechanism achieves the highest scores in both safety and comfort metrics, leading to the best overall performance.
Further, compared across blocks under the same $8\,\mathrm{s}$ setting, pivot-conditioned diffusion mechanism with Contingency Score surpasses the \textsc{PDM} baseline and also slightly edges the $4\,\mathrm{s}$ \textsc{PDM} setup overall, chiefly by improving comfort and TTC. This indicates that our method better balances near-term executability and long-term preparedness.

\section{Conclusion and Future Work}

In this work, we present \textit{CoPlanner}, a unified generation--evaluation framework that couples short-term conditioned diffusion with contingency-aware scoring. By enforcing a safety-validated shared short-term segment and sampling diverse long-term branches, CoPlanner stabilizes near-term execution while retaining feasible fallback options when interactions evolve.
On the nuPlan closed-loop benchmark, CoPlanner delivers consistent gains: it attains state-of-the-art performance on \textbf{Val14} in both non-reactive and reactive modes and on \textbf{Test14} (reactive), and remains competitive on the other splits. Ablations indicate that (i) the two-stage generation with a shared short-term segment and (ii) the multi-scenario aggregation in the evaluator are the primary drivers of the improvement, yielding better safety and comfort without sacrificing progress.


\noindent\textbf{Limitations and Future Work.}
CoPlanner may still fail when all sampled futures are infeasible, and the current aggregator does not model scenario likelihoods or calibrated uncertainty. The branching time $t_b$ is fixed, which can be suboptimal across scenes, and the computational budget for multi-branch evaluation can be further reduced. 
Future work will add feasibility-aware guidance during denoising, incorporate physics-informed and dynamics-informed priors, and adaptively choose the branching horizon. It will also aggregate scenarios using likelihood or risk weights under calibrated uncertainty, and broaden closed-loop evaluations with stronger reactive agents and real-world tests.

\bibliography{sample}
\bibliographystyle{ieeetr}
\end{document}